\documentclass[fleqn,10pt]{wlscirep}

\usepackage[utf8]{inputenc}
\usepackage[T1]{fontenc}
\usepackage{graphicx}
\usepackage{amsmath}
\usepackage{url}
\usepackage{algorithm}
\usepackage{algpseudocode}
\usepackage{adjustbox}
\usepackage[switch]{lineno}
\usepackage{multirow} 
\usepackage{diagbox}
\usepackage{float}
\usepackage{subcaption}
\usepackage{soul}
\usepackage{listings}
\usepackage{arydshln}

\newcommand{\impn}{\textsc{I-MPN}}

\title{\impn: Inductive Message Passing Network for Efficient Human-in-the-Loop Annotation of Mobile Eye Tracking Data}

\author[+,1,2,3,*]{ Hoang H. Le}
\author[+,1,4,5,*]{Duy M. H. Nguyen}
\author[1]{Omair Shahzad Bhatti}
\author[1]{László Kopácsi}
\author[2]{Thinh P. Ngo}
\author[2]{Binh T. Nguyen}
\author[1,6]{Michael Barz}
\author[1,6]{Daniel Sonntag}
\affil[1]{German Research Center for Artificial Intelligence (DFKI), Interactive Machine Learning Department,  66123 Saarbrücken, Germany}
\affil[2]{University of Science, VNU-HCM, Mathematics and Computer Science Department, Ho Chi Minh City, Vietnam}
\affil[3]{Quy Nhon AI Research and Development Center, FPT Software, Vietnam}
\affil[4]{Max Planck Research School for Intelligent Systems (IMPRS-IS), 70569 Stuttgart, Germany}
\affil[5]{Univerity of Stuttgart, Machine Learning and Simulation Science Department, 70569 Stuttgart, Germany}
\affil[6]{University of Oldenburg, Applied Artificial Intelligence Department, 26129 Oldenburg, Germany}
\affil[*]{Corresponding author  ho\_minh\_duy.nguyen@dfki.de}

\affil[+]{these authors contributed equally to this work.}


\begin{abstract}
Comprehending how humans process visual information in dynamic settings is crucial for psychology and designing user-centered interactions. While mobile eye-tracking systems combining egocentric video and gaze signals can offer valuable insights, manual analysis of these recordings is time-intensive. In this work, we present a novel \textit{human-centered learning algorithm} designed for automated object recognition within mobile eye-tracking settings. Our approach seamlessly integrates an object detector with a spatial relation-aware inductive message-passing network (I-MPN), harnessing node profile information and capturing object correlations. Such mechanisms enable us to learn embedding functions capable of generalizing to new object angle views, facilitating rapid adaptation and efficient reasoning in dynamic contexts as users navigate their environment.
Through experiments conducted on three distinct video sequences, our \textit{interactive-based method} showcases significant performance improvements over fixed training/testing algorithms, even when trained on considerably smaller annotated samples collected through user feedback. Furthermore, we demonstrate exceptional efficiency in data annotation processes and surpass prior interactive methods that use complete object detectors, combine detectors with convolutional networks, or employ interactive video segmentation.
\end{abstract}
\begin{document}

\flushbottom
\maketitle

\thispagestyle{empty}

\section{Introduction}
The advent of mobile eye-tracking technology has significantly expanded the horizons of research in fields such as psychology, marketing, and user interface design by providing a granular view of user visual attention in naturalistic settings \cite{holmqvist2011eye,duchowski2017eye}. This technology captures details of eye movement, offering insights into cognitive processes and user behavior in real-time scenarios such as interacting with physical products or mobile devices. However, the manual analysis of eye-tracking data is challenging due to the extensive volume of data generated and the complexity of dynamic visual environments where target objects may overlap and be affected by environmental noise \cite{strandvall2009eye,gardony2020eye}. These barriers underscore the necessity for autonomous analytical strategies, leveraging computational algorithms to streamline data processing and mitigate human error. 

\begin{figure}
    \centering
    \includegraphics[width=.8\textwidth]{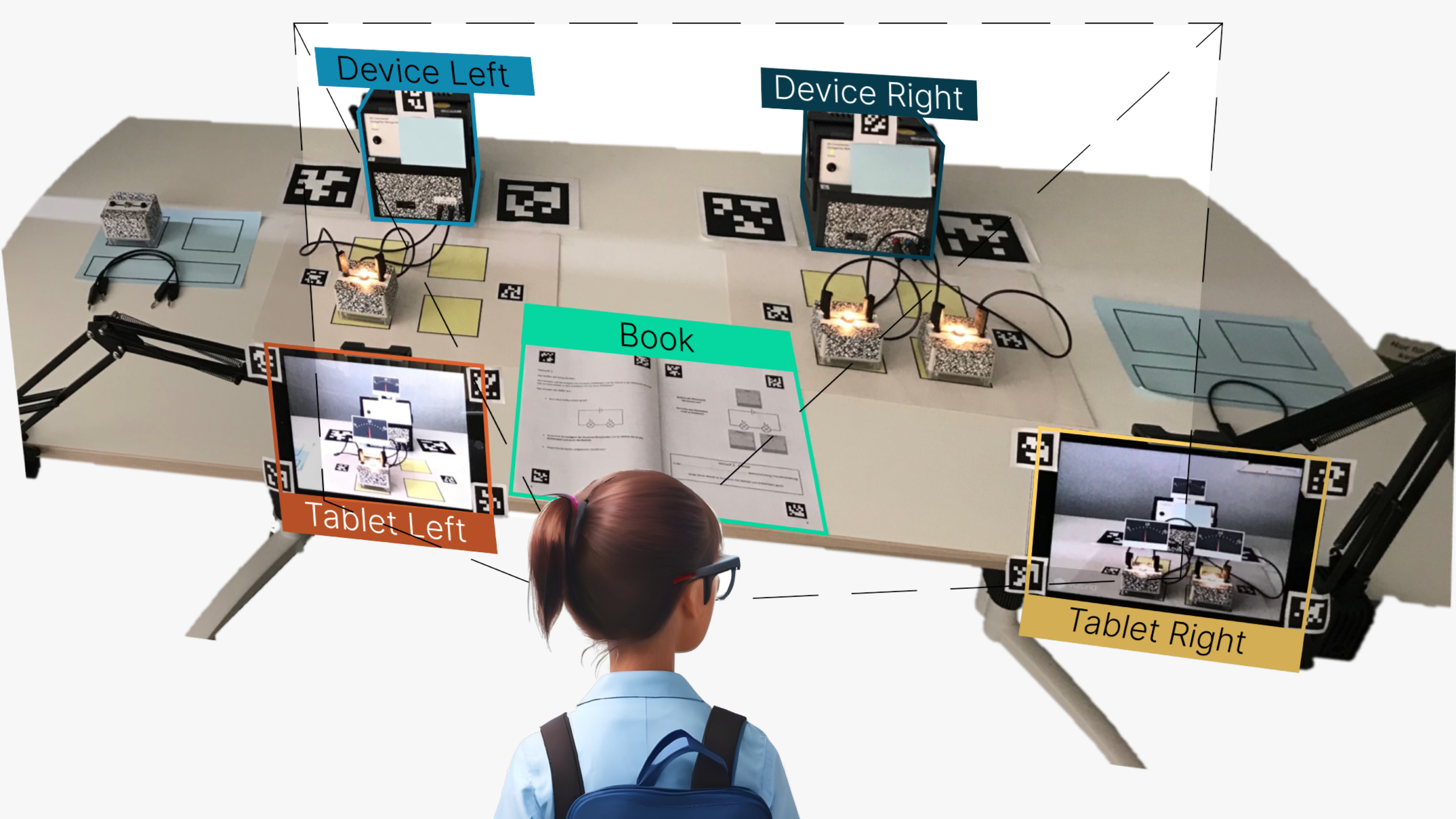}
    \caption{Our mobile eye-tracking setup with different viewpoints.}
    \label{fig:environment}
\end{figure}

To this end, machine learning methods have been extensively applied 
across various domains, including gaze estimation, area of interest detection, and visual attention detection. Notably, models utilizing convolutional neural networks (CNNs), recurrent neural networks (RNNs), and object detection are proposed to achieve high accuracy and efficiency in these tasks
\cite{zhang2017mpiigaze,yang2020siamatt,barz2021automatic}. Nonetheless, these approaches usually encounter substantial challenges rooted in the human factor. Foremost, the dynamic nature of eye movements across users and contexts \cite{wei2018and,hu2021ehtask} causes models to be sensitive to occlusions and illumination, requiring large annotated data to maintain accuracy. Additionally, integrating user feedback into the learning process remains problematic \cite{wu2022survey} where models are required to pay attention to individual preferences and situational context, which is crucial for improving the usability and effectiveness of mobile eye-tracking systems.

In this study, we present a new approach aimed at enhancing object recognition under interactive mobile eye-tracking (Figure \ref{fig:environment}), specifically optimizing data annotation efficiency and advancing human-in-the-loop learning models (Figure \ref{fig:overview_method}). Equipped with eye-tracking devices, users generate video streams alongside fixation points, providing visual focus as they navigate through their environment. Our primary aim lies in recognizing specific objects, such as tablet-left, tablet-right, book, device-left, and device-right, with all other elements considered background, as demonstrated in Figure~\ref{fig:environment}. To kickstart the training process with initial data annotations, we leverage video object segmentation (VoS) techniques \cite{wang2021swiftnet,cheng2022xmem}. Users are prompted to provide weak scribbles denoting areas of interest (AoI) and assign corresponding labels in initial frames. Subsequently, the VoS tool autonomously extrapolates segmentation boundaries closest to the scribbled regions, thereby generating predictions for later frames. During a period of time, users interact with the interface, reviewing and refining results by manipulating scribbles or area-of-effect (AoE) labels if they reveal error annotations.

In the next phase, we collect segmentation masks and correspondence annotations provided by the VoS tool to define bounding boxes encompassing AoI and their corresponding labels to train recognition algorithms. 
Our approach, named I-MPN, consists of two primary components: (i) an object detector tasked with generating proposal candidates within environmental setups and (ii) an \underline{I}nductive \underline{M}essage-\underline{P}assing \underline{N}etwork \cite{hamilton2017inductive,ciano2021inductive,qu2021neural} designed to discern object relationships and spatial configurations, thereby determining the labels of objects present in the current frame based on their correlations. It is crucial to highlight that identical objects may bear different labels contingent upon their spatial orientations (e.g., left, right) in our settings (Figure \ref{fig:environment}, device left and right). This characteristic often poses challenges for methods reliant on local feature discrimination, such as object detection or convolutional neural networks, due to their inherent lack of global spatial context. I-MPN, instead, can overcome this issue by dynamically formulating graph structures at different frames whose node features are represented by bounding box coordinates and semantic feature representations inside detected boxes derived from the object detector. Nodes then exchange information with their local neighborhoods through a set of trainable aggregator functions, which remain invariant to input permutations and are adaptable to unseen nodes in subsequent frames. Through this mechanism, I-MPN plausibly captures the intricate relationships between objects, thus augmenting its representational capacity to dynamic environmental shifts induced by user movement.

Given the initial trained models, we integrate them into a human-in-the-loop phase to predict outcomes for each frame in a video. If users identify erroneous predictions, they have the ability to refine the models by providing feedback through drawing scribbles on the current frame using VoS tools, as shown in Figure~\ref{fig:annotation_tool}. This feedback triggers the generation of updated annotations for subsequent frames, facilitating a rapid refinement process similar to the initial annotation stage but with a reduced timeframe. The new annotations are then gathered and used to retrain both the object detector and message-passing network in the backend before being deployed for continued inference. If errors persist, the iterative process continues until the models converge to produce satisfactory results. We illustrate such an iterative loop in Figure~\ref{fig:overview_method}. 

In summary, we observe the following points: 
\begin{itemize}
    \item Firstly, I-MPN proves to be highly efficient in adapting to user feedback within mobile eye-tracking applications. Despite utilizing a relatively small amount of user feedback data ($20\%-30\%$), we achieve 
    performance levels that are comparable to or even exceed those of conventional methods, which typically depend on fixed training data splitting rates of $70\%$.
    \item Secondly, a comparative analysis with other human-learning approaches, such as object detectors and interactive segmentation methods, highlights the superior performance of I-MPN, especially in dynamic environments influenced by user movement. This underscores I-MPN's capability to comprehend object relationships in challenging conditions.
    \item Finally, we measure the average user engagement time needed for initial model training data provision and subsequent feedback updates. Through empirical evaluation of popular annotation tools in segmentation and object classification, we demonstrate I-MPN's time efficiency, reducing label generation time by $60\%-70\%$. We also investigate factors influencing performance, such as message-passing models. Our findings confirm the adaptability of the proposed framework across diverse network architectures.
\end{itemize}

\section{Related Work}

\subsection{Eye tracking-related machine learning models}
Many mobile eye-tracking methods rely on pre-trained computer vision models. For example, some methods automatically map fixations to bounding boxes using pre-trained object detection models \cite{venuprasad_analyzing_2020,deane_deep-saga_2022}
, while others classify image patches around fixation points using pre-trained image classification models \cite{barz2021automatic}. However, these approaches are typically confined to highly constrained settings where the training data aligns with the target domain. Studies have revealed substantial discrepancies between manual and automatic annotations for areas of interest (AOIs) corresponding to classes in benchmark datasets like COCO \cite{lin2014microsoft}, highlighting challenges in adapting pre-trained models to realistic scenarios with diverse domains \cite{deane_deep-saga_2022}. 
Alternative strategies involve fine-tuning object detection models for specific target domains \cite{batliner2020automated,kumari2021mobile}, but these lack interactivity during training and cannot dynamically adjust models during annotation. While some interactive methods for semi-automatic data annotation exist, they often rely on non-learnable feature descriptions such as color histograms or bag-of-SIFT features \cite{kurzhals2016visual,panetta2019software}. Recently Kurzhals et al. \cite{kurzhals2020visual} introduced an interactive approach for annotating and interpreting egocentric eye-tracking data for activity and behavior analysis, utilizing iterative time sequence searches based on eye movements and visual features. However, their method annotates objects by cropping image patches around each point of gaze, segmenting the patches, and presenting representative gaze thumbnails as image clusters on a 2D plane. Unlike these works, our I-MPN is designed 
to capture both \textit{local visual feature} representations and \textit{global interactions} among objects by inductive message passing network, making models robust under occluded or vastly change point of view conditions. 

\subsection{Graph neural networks for object recognition}
Graph neural networks (GNNs) are neural models designed for analyzing graph-structured data like social networks, biological networks, and knowledge graphs \cite{zhou2020graph}. Beyond these domains, GNNs can be applied in object recognition to identify and locate objects in images or videos by leveraging graph structures to encode spatial and semantic relations among objects or regions. Through mechanisms like graph convolution \cite{Kipf:2016tc} or attention mechanisms \cite{velikovi2017graph}, GNNs efficiently aggregate and propagate information across the graph. Notable methods employing GNNs for object recognition include KGN \cite{liu2020od}, SGRN \cite{xu2019spatial}, and RGRN \cite{zhao2023rgrn}, among others. 
However, in mobile eye-tracking scenarios, these methods face two significant challenges. Firstly, the message-passing mechanism typically operates on the entire graph structure, necessitating a fixed set of objects during both training and inference. This rigidity implies that the entire model must be updated to accommodate new, unseen objects that may arise later due to user interests. Secondly, certain methods, such as RGRN \cite{zhao2023rgrn}, rely on estimating the co-occurrence of pairs of objects in scenes based on training data, yet such information is not readily available in human-in-the-loop settings where users only provide small annotated samples, resulting in co-occurrence matrices among objects evolve over time. I-MPN tackles these issues by performing message passing to aggregate information from neighboring nodes, enabling the model to maintain robustness to variability in the graph structure across different instances. While there exist works have exploited this idea for link predictions \cite{hamilton2017inductive}, recommendation systems \cite{graphsaint-iclr20}, or video tracking \cite{prummel2023inductive}, we the first propose a formulation for human interaction in eye-tracking setups.

\section{Methodology}

\begin{figure}
    \centering
    \includegraphics[width=1.0\textwidth]{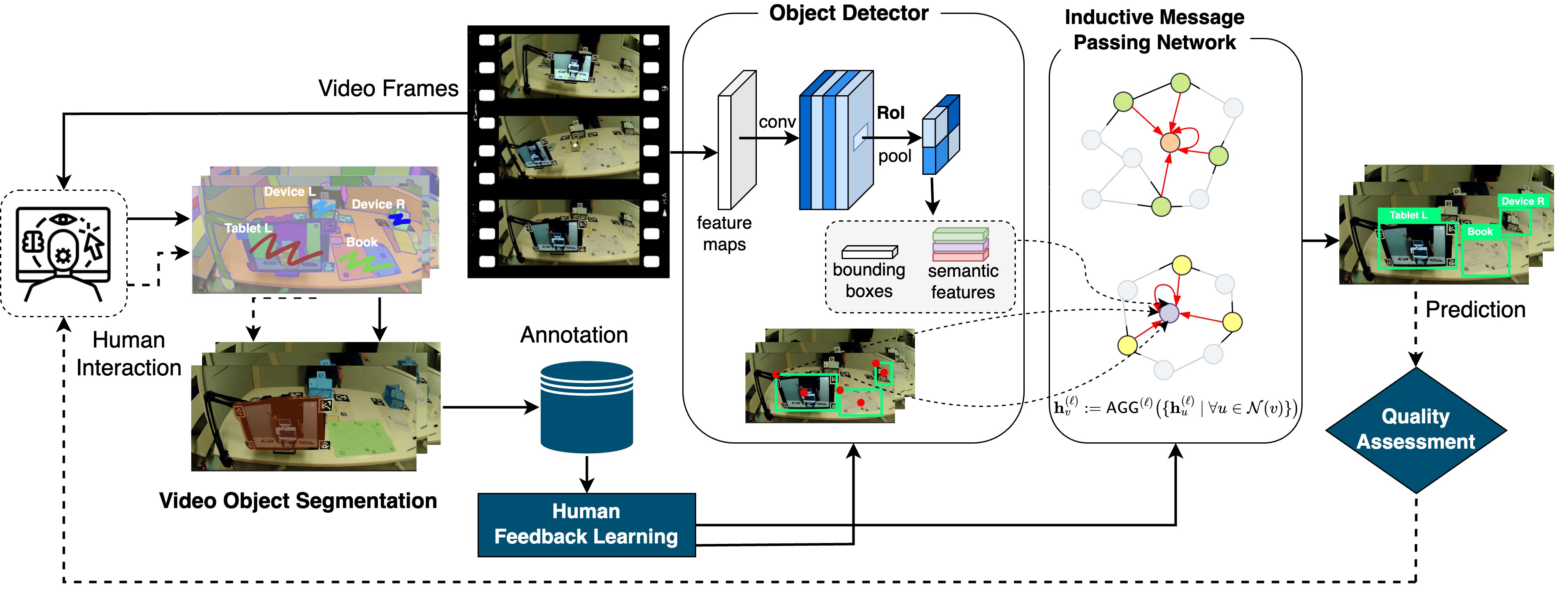}
    \caption{Overview our human-in-the-loop I-MPN approach. The bottom dashed arrow indicates the feedback loop. The human interacts with the video object segmentation algorithm to generate annotations used to train an object detector and another graph reasoning network.}
    \label{fig:overview_method}
\end{figure}

\subsection{Overview Systems}
Figure \ref{fig:overview_method} illustrates the main steps in our pipeline. Given a set of video frames: (i) the user generates annotations by scribbling or drawing boxes around objects of interest, which are then fed into the video object segmentation algorithm to generate segment masks over the time frames. (ii) The outputs are subsequently added to the database to train an object detector, perform spatial reasoning, and generate labels for appearing objects using inductive message-passing mechanisms. The trained models are then utilized to infer the next frames until the user interrupts upon encountering incorrect predictions. At this point, users provide feedback as in step (i) for these frames (Figure~\ref{fig:overview_method} bottom dashed arrow). New annotations are then added to the database, and the models are retrained as in step (ii). This loop is repeated for several rounds until the model achieves satisfactory performance.
In the following sections, we describe our efficient strategy for enabling users to quickly generate annotations for video frames (Section~\ref{sec:vos}) and our robust machine learning models designed to quickly adapt from user feedback to recognize objects in dynamic environments (Section~\ref{sec:detector-gnn}).

\subsection{User Feedback as Video Object Segmentation}\label{sec:vos}
Annotating objects in video on a frame-by-frame level presents a considerable time and labor investment, particularly in lengthy videos containing numerous objects. To surmount these challenges, we utilize video object segmentation-based methods \cite{yao2020video,zhou2022survey}, significantly diminishing the manual workload. With these algorithms, users simply mark points or scribble within the Area of Interest (AoI) along with their corresponding labels (Figure \ref{fig:annotation_tool}). Subsequently, the VoS component infers segmentation masks for successive frames by leveraging spatial-temporal correlations (Figure \ref{fig:overview_method}-left). These annotations are then subject to user verification and, if needed, adjustments, streamlining the process rather than starting from scratch each time.

Particular, VoS aims to identify and segment objects across video frames (\( \{F_1, F_2, \ldots, F_T\} \)), producing a segmentation mask \( M_t \) for each frame \( F_t \). We follow \cite{cheng2022xmem} to apply a cross-video memory mechanism to maintain instance consistency, even with occlusions and appearance changes. In the first step, for each frame \( F_t \), the model extracts a set of feature vectors \( \mathbf{F}_t = \{f_{t1}, f_{t2}, \ldots, f_{tn}\} \), where each \( f_{ti} \) corresponds to a region proposal in the frame and \( n \) is the total number of proposals. Another \textit{memory module} maintains a memory $ \mathbf{M}_t = \{m_1, m_2, \ldots, m_k\} $ that stores aggregated feature representations of previously identified object instances, where \( k \) is the number of unique instances stored up to frame \( F_t \). To generate correlation scores $\displaystyle \mathbf{C}_t = \{c_{t1}, c_{t2}, \ldots, c_{tn}\}$ among consecutive frames, a \textit{memory reading function} $ \displaystyle \mathbf{R}(\mathbf{F}_t, \mathbf{M}_{t-1}) \rightarrow \mathbf{C}_t$ 
is used. The scores in $\mathbf{C}_{t}$ estimate the likelihood of each region proposal in \( F_t \) matching an existing object instance in memory. The memory is then updated via a writing function $\mathbf{W}(\mathbf{F}_t, \mathbf{M}_{t-1}, \mathbf{C}_t) \rightarrow \mathbf{M}_t$, which modifies \( \mathbf{M}_t \) based on the current observations and their correlations to stored instances. Finally, given the updated memory and correlation scores, the model assigns to each pixel in frame $\mathbf{F}_t$ a label and an instance ID, represented by $\mathbf{S}(\mathbf{F}_t, \mathbf{M}_t, \mathbf{C}_t) \rightarrow \{(l_{t1}, i_{t1}), (l_{t2}, i_{t2}), \ldots, (l_{tn}, i_{tn})\}$, where $ (l_{ti}, i_{ti})$ indicates the class label and instance ID for the $i$-th proposal.

By using cross-video memory, the method achieved promising accuracy in various tasks ranging from video understanding \cite{song2023moviechat}, robotic manipulation \cite{huang2023voxposer}, or neural rendering \cite{tschernezki2024epic}. In this study, we harness this capability as an efficient tool for user interaction in annotation tasks, particularly within mobile eye-tracking, facilitating learning and model update phases.
The advantages of used VoS over other prevalent annotation methods in segmentation are presented in Table \ref{tab:annotation_compare_time}.

\subsection{Dynamic Spatial-Temporal Object Recognition}
\label{sec:detector-gnn}
\subsubsection*{Generating Candidate Proposals}
Due to the powerful learning ability of deep convolutional neural networks, object detectors such as Faster R-CNN \cite{girshick2015fast} and YOLO \cite{redmon2016you,jiang2022review} offer high accuracy, end-to-end learning, adaptability to diverse scenes, scalability, and real-time performance. However, they still only propagate the visual features of the objects within the region proposal and ignore complex topologies between objects, leading to difficulties distinguishing difficult samples in complex spaces.
Rather than purely using object detector outputs, we leverage their bounding boxes and corresponding semantic feature maps at each frame as candidate proposals, which are then inferred by another relational graph network. In particular, denoting $\mathbf{f}_{\theta}$ as the detector, at the $i$-th frame $F_{i}$, we compute a set of $k$ bonding boxes cover AoE regions by $\mathbf{B}_{i} = \{b_{i1},b_{i2},...,b_{ik}\}$ and feature embeddings inside those ones $\mathbf{Z}_{i} = \{z_{i1},z_{i2},...,z_{ik}\}$ while ignoring $\mathbf{P}_{i}$ denotes the set of class probabilities for each bounding boxes in $\mathbf{B}_{i}$ where $\{\mathbf{B}_{i}, \mathbf{Z}_{i}, \mathbf{P}_{i}\} \leftarrow \mathbf{f}_{\theta}(F_{i})$. The $\mathbf{f}_{\theta}$ is trained and updated with user feedback with annotations generated from the VoS tool.

\renewcommand{\Comment}[2][0.65\linewidth]{%
  \leavevmode\hfill\makebox[#1][l]{//~#2}}
  
\begin{algorithm}
\caption{\texttt{I-MPN Forward and Backward Pass}}
\begin{algorithmic}[1]
\State \textbf{Input:} Graph $G(V,E)$, input features $\{x_v \in X, \forall v \in V\}$, \\
depth $K$, weight matrices $\{W^{(k)}, \forall k = 1...K\}$, non-linearity $\sigma$, \\
differentiable aggregator functions AGGREGATE$_k$, \\
neighborhood function $N: V \rightarrow 2^V$
\State \textbf{Output:} Vector representations $z_v$ for all $v \in V$
\Procedure{I-MPN\,Forward}{${G}, {X}, K$}
    \For{$k = 1$ \textbf{to} $K$}
        \For{\textbf{each} node $v \in V$}
            \State $h_{N(v)}^{(k)} \leftarrow \mathsf{AGG}_k(\{h_u^{(k-1)}, \forall u \in N(v)\})$
            \State $h_v^{(k)} \leftarrow \sigma\left(W^{(k)} \cdot \mathsf{CONCAT}(h_v^{(k-1)}, h_{N(v)}^{(k)})\right)$
        \EndFor
    \EndFor
        \For{each node $v \in V$}
        \State $\hat{y}_v \gets \mathsf{SOFTMAX}\left(W^{o} \cdot h_v^{(K)}\right)$ \Comment{predictions for each node}
    \EndFor
    \State $\mathcal{L} \gets -\sum_{v \in V} \sum_{c=1}^{C} Y_{v,c} \log(\hat{y}_{v,c})$ \Comment{compute cross-entropy loss}
    \State \Return $L$
\EndProcedure
\State
\Procedure{I-MPN\,Backward}{$\mathcal{L}, W$}
    \For{$k = K$ \textbf{down to} $1$}
        \State Compute gradients: $\frac{\partial \mathcal{L}}{\partial W^{(k)}}$ using chain rule
        \State Update weights: $W^{(k)} \leftarrow W^{(k)} - \eta \frac{\partial \mathcal{L}}{\partial W^{(k)}}$
    \EndFor
\EndProcedure
\end{algorithmic}
\label{algo:forward-backward}
\end{algorithm}

\subsection*{Inductive Message Passing Network}
We propose a graph neural network $\mathbf{g}_{\epsilon}$ using 
inductive message-passing operations \cite{hamilton2017inductive,ciano2021inductive} for reasoning relations among objects detected within each frame in the video. Let $\mathbf{G}_{i} = (\mathbf{V}_{i}, \mathbf{E}_{i})$ denote the graph at the $i$-th frame where $\mathbf{V}_{i}$ being nodes with each node $v_{ij} \leftarrow b_{ij} \in\,\mathbf{V}_{i}$ defined from bounding boxes $\mathbf{B}_{i}$. $\mathbf{E}$ is the set of edges where we permit each node to be fully connected to the remaining nodes in the graph. We initialize node-feature matrix $\mathbf{X}_{i}$, which associates for each  $v_{ij} \in V_{i}$ a feature embedding $x_{v_{ij}}$. In our setting, we directly use $x_{v_{ij}} = z_{ij} \in Z_{i}$ taken from the output of the object detector.
Most current GNN approaches for object recognition \cite{xu2019spatial,zhao2023rgrn} use the following framework to compute feature embedding for each node in the input graph $\mathbf{G}$ (for the sake of simplicity, we ignore frame index):
\begin{equation}
\mathbf{H}^{(l+1)} = \sigma(\tilde{D}^{-\frac{1}{2}} \tilde{\mathbf{A}} \tilde{D}^{-\frac{1}{2}} \mathbf{H}^{(l)} \mathbf{W}^{(l)})
\label{eq:gcn}
\end{equation}
where: $\mathbf{H}^{(l)}$ represents all node features at layer \(l\), $\tilde{\mathbf{A}}$ is the adjacency matrix of the graph $\mathbf{G}$ with added self-connections, $\tilde{D}$ is the degree matrix of $\tilde{\mathbf{A}}$, $\mathbf{W}^{(l)}$ is the learnable weight matrix at layer $l$, $\sigma$ is the activation function,  $\mathbf{H}^{(l+1)}$ is the output node features at layer $l+1$. To integrate prior knowledge,  Zhao, Jianjun, et al. \cite{zhao2023rgrn} further counted co-occurrence between objects as the adjacency matrix $\tilde{\mathbf{A}}$. However, because the adjacency matrix $\tilde{\mathbf{A}}$ is fixed during the training, \textit{the message passing operation in Eq\,\eqref{eq:gcn} cannot generate predictions for new nodes that were not part of the training data appear during inference}, i.e., the set of objects in the training and inference has to be identical. This obstacle makes the model unsuitable for the mobile eye-tracking setting, where users' areas of interest may vary over time. 
We address such problems by changing the way node features are updated, from being dependent on the entire graph structure  $\tilde{\mathbf{A}}$ to neighboring nodes $\mathcal{N}(v)$ for each node $v$. In particular, 
\begin{equation}\centering
    \mathbf{h}_{\mathcal{N}(v)}^{(l)} = \mathsf{AGG}^{(\ell)}(\{\mathbf{h}_u^{(l)}, \forall u \in \mathcal{N}(v)\})
\end{equation}
\begin{equation}
\mathbf{h}_v^{(l+1)} = \sigma\big(\mathbf{W}^{(l)} \cdot \mathsf{CONCAT}\big(\mathbf{h}_v^{(l)}, \mathbf{h}_{\mathcal{N}(v)}^{(l)}\big)\big)
\end{equation}
where: $\mathbf{h}_v^{(l)}$ represents the feature vector of node $v$ at layer $l$,
$\mathsf{AGG}$ is an aggregation function (e.g., Pooling, LSTM), $\mathsf{CONCAT}$ be the concatenation operation, $\mathbf{h}_v^{(l+1)}$ is the updated feature vector of node $v$ at layer $l+1$. 
In scenarios when a new unseen object $v_{new}$ is added to track by the user, we can aggregate information from neighboring seen nodes $v_{seen} \in \mathcal{N}(v_{new})$ by:
\begin{equation}
    \mathbf{h}_{v_{new}}^{(l+1)} = \sigma\big(\mathbf{W}^{(l)} \cdot \mathsf{CONCAT}\big(\mathbf{h}_{v_{new}}^{(l)}, \mathsf{AGG}^{(\ell)}(\{\mathbf{h}_{v_{seen}}^{(l)}\})
\end{equation}
and then update the trained model on this new sample rather than all nodes in training data as Eq.\eqref{eq:gcn}. The forward and backward pass of our message-passing algorithm is summarized in the Algorithm \ref{algo:forward-backward}.
We found that such operations obtained better results in experiments than other message-passing methods such as attention network \cite{velikovi2017graph}, principled aggregation \cite{corso2020principal} or transformer \cite{shi2020masked} (Figure \ref{fig:gnn_models}). 

\begin{algorithm}[H]
   \caption{PyTorch-style I-MLE algorithm.}
   \label{alg:i-mle}
   
    \definecolor{codeblue}{rgb}{0.25,0.5,0.5}
    \lstset{
      basicstyle=\fontsize{8pt}{8pt}\ttfamily\bfseries,
      commentstyle=\fontsize{8pt}{8pt}\color{codeblue},
      keywordstyle=\fontsize{8pt}{8pt},
    }
\begin{lstlisting}[language=python]
1:  # f_theta: object detector
2:  # g_epsilon: inductive message passing network
3:  # max_update: maximum number of taking user feedback
4:  # VoS: video object segmentation model
5:  # t_initial: time for initial annotation step
6:  # t_update: time for updating with user feedback
7:  # F = [F_1, ..., F_t]: list of frames in video

## Stage 1. Training initial models
    # extract initial annotations by user (Alg. 3)
8:  D_init = interactive_func(F[0:t_initial], VoS)
    # train object detector and relational graph network
9:  f_theta.train(D_init); g_epsilon.train(D_init)

## Stage 2. Inference and User Feedback Update
10: update_time = 0
11: frame_index = t_initial
12: while frame_index <= len(F) + 1:
       # generate object candidates by the detector
13:    candidate_objects, feature_maps = f_theta(F[frame_index])

       # build graph and inference labels
15:    G = construct_graph(candidate_objects, feature_maps)
16:    detected_objects, labels = g_epsilon(G)

       # show outputs to user
17:    display(detected_objects, labels) 

       # user feedback if encountering wrong outputs
18:    if (update_time <= max_update) and (user.satisfy(detected_objects, label) is False):
19:        start_index = frame_index
20:        end_index = start_index + t_update + 1

           # using Alg. 3
21:        D_feedback = interactive_func(F[start_index, end_index], VoS)

           # updated model with user feedback
22:        f_theta.train(D_feedback); 
23:        g_epsilon.train(D_feedback)

           # update counting numbers
24:        update_time += 1 
25:        frame_index = end_index            
26:    else:
27:        frame_index += 1
\end{lstlisting}
\end{algorithm}

\subsubsection*{End-to-end learning from Human Feedback}
In Algorithm \ref{alg:i-mle}, we present the proposed human-in-the-loop method for mobile eye-tracking object recognition. This approach integrates user feedback to jointly train the object detector $\mathbf{f}_{\theta}$ and the graph neural network $\mathbf{g}_{\epsilon}$ for spatial reasoning of object positions. Specifically, $\mathbf{f}_{\theta}$ is trained to generate coordinates for proposal object bounding boxes, which are then used as inputs for $\mathbf{g}_{\epsilon}$ (bounding box coordinates and feature embedding inside those regions). The graph neural network $\mathbf{g}_{\epsilon}$ is, on the other hand, trained to generate labels for these objects by considering the correlations among them. Notably, our pipeline operates as an end-to-end framework, optimizing both the object detector and the graph neural network simultaneously rather than as separate components. This lessens the propagation of errors from the object detector to the GNN component, making the system be robust to noises in environment setups.  
The trained models are deployed afterward to infer the next frames and are then refined again at wrong predictions, giving user annotation feedback in a few loops till the model converges. In the experiment results, we found that such a human-in-the-loop scheme enhances the algorithm's adaptation ability and 
yields comparable or superior results to traditional learning methods with a set number of training and testing samples.

\begin{algorithm}
   \caption{User feedback propagation algorithm}
   \label{alg:user_feedback}
   
    \definecolor{codeblue}{rgb}{0.25,0.5,0.5}
    \lstset{
      basicstyle=\fontsize{8pt}{8pt}\ttfamily\bfseries,
      commentstyle=\fontsize{8pt}{8pt}\color{codeblue},
      keywordstyle=\fontsize{8pt}{8pt},
    }
    \begin{lstlisting}[language=python]
## User feedback functions            
1: def interactive_func(list_frame, VoS):
2:    D = [] # store annotation data
    
      # generate initial segment masks
3:    init_mask = VoS(list_frames[0]) 
4:    display(init_mask)
    
      # user correct with scribbles
5:    ann_mask, label = user.annotate(init_mask)

      # propagate predictions for next frames
6:    for frame in sorted(list_frames[1:]):
7:        next_mask, label = VoS(frame, ann_mask, label)
8:        display(next_mask, label)
        
          # user update if persist errors
9:        if user.satisfy(updated_mask, label) is False:
10:            ann_mask, label = user.annotate(next_mask, label)
11:            D.append({ann_mask, label, frame})
12:        else:
13:            D.append({next_mask, label, frame})  
14:    return D    
\end{lstlisting}
\end{algorithm}

\begin{figure}
    \centering
    \includegraphics[width=\textwidth]{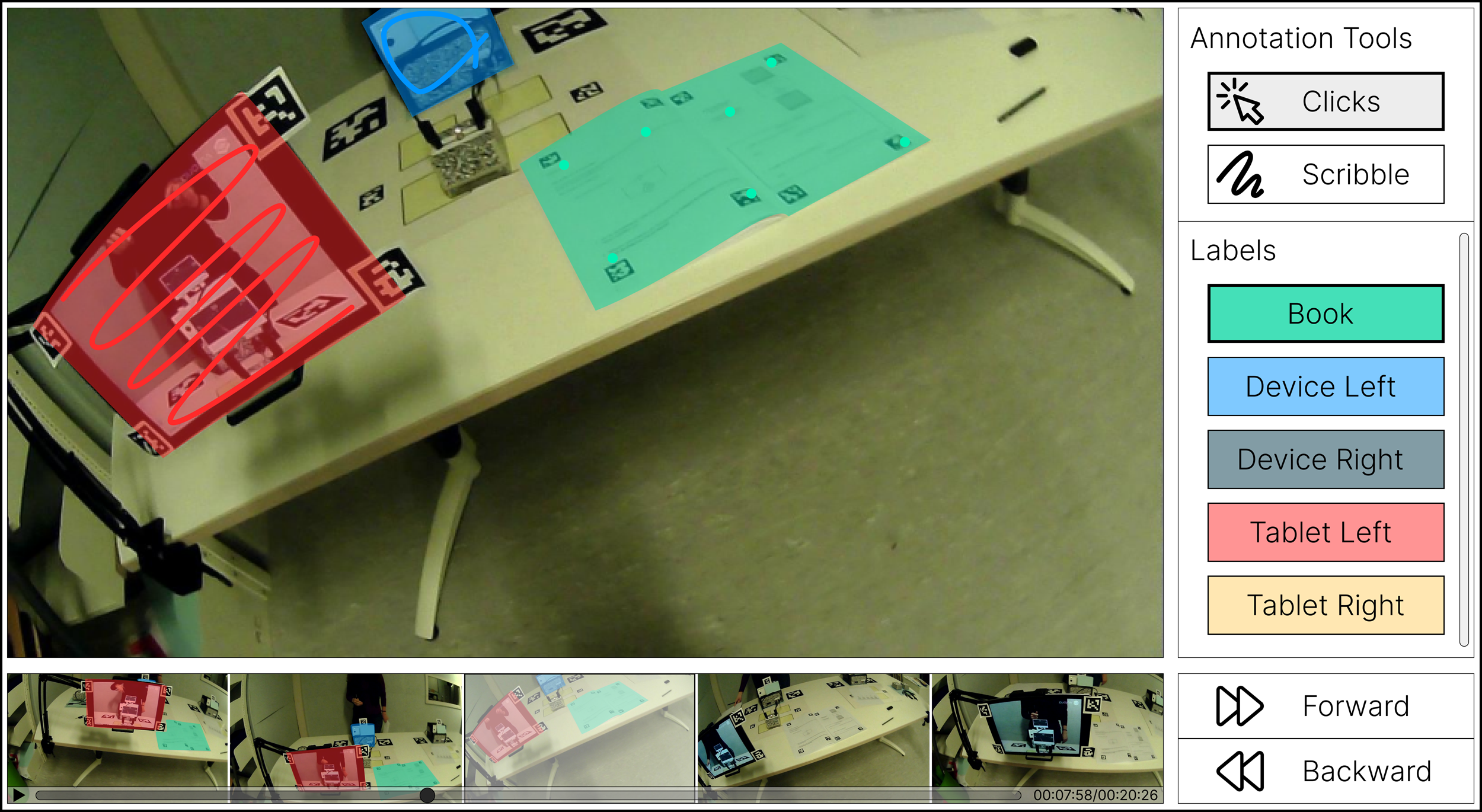}
    \caption{The video object segmentation-based interface allows users to annotate frames using weak prompts like clicks and scribbles, then propagate these annotations to subsequent frames.}
    \label{fig:annotation_tool}
\end{figure}

\section{Experiments \& Results}
\subsection{Dataset}
Figure \ref{fig:environment} illustrates our experimental setup where we record three video sequences captured by different users, each occurring in two to three minutes (Table \ref{tab:annotation_compare_time}). The users wear an eye tracker on their forehead, which records what they observe over time while also providing fixation points, showing the user's focus points at each time frame. We are interested in detecting five objects: tables (left, right), books, and devices (left, right).

\paragraph{Video Ground-Truth Annotations} To generate data for model evaluation, we asked users to annotate objects in each video frame using the VoS tool introduced in Section~\ref{sec:vos}. Following the cross-entropy memory method as described in \cite{cheng2022xmem}, we interacted with users by displaying segmentation results on a monitor. Users then labeled data and created ground truths by clicking the "Scribble" and "Adding Labels" functions for objects. Subsequently, by clicking the "Forward" button, the VoS tool automatically segmented the objects' masks in the next frames until the end of the video.  If users encountered incorrectly generated annotations, they could click "Stop" to edit the results using the "Scribble" and "Adding Labels" functions again (Figure \ref{fig:annotation_tool}). Table \ref{tab:annotation_compare_time} highlights the advantages of the VoS method for video annotation compared to popular tools used in object detection or semantic segmentation.

\paragraph{Metrics}  The experiment results are measured by the consistency of predicted bounding boxes and their labels with ground-truth ones. 
In most experiments except the fixation point cases, we evaluate performance for all objects in each video frame. We define $AP@\alpha$ as the Area Under the Precision-Recall Curve (AUC-PR) evaluated at $\alpha$ IoU threshold 
$\displaystyle AP@\alpha =\int ^{1}_{0}p\left( r\right) dr$ where $p(r)$ represents the precision at a given recall level $r$. 
The mean Average Precision \cite{everingham2010pascal} is computed at different $\alpha$ IoU ($mAP@\alpha$), which is the average of AP values over all classes, i.e., $ \displaystyle mAP@\alpha =\frac{1}{n}\sum ^{n}_{i=1}(AP@\alpha)_{i}$. We provide results for $\alpha \in \{50, 75\}$. Furthermore, we report $mAP$ as an average of different IoU ranging from $0.5 \rightarrow 0.95$ with a step of $0.05$.

\paragraph{Model Configurations} We use the Faster-RCNN \cite{girshick2015fast} as the network backbone for the object detector $\mathbf{f_{\theta}}$ and follow the same proposed training procedure by the authors. The message-passing component $\mathbf{g_{\epsilon}}$ uses the $\mathsf{MaxPooling}$ and $\mathsf{LSTM}$ aggregator functions to extract and learn embedding features for each node.  We use output bounding boxes and feature embedding at the last layer in $\mathbf{f_{\theta}}$ as inputs for $\mathbf{g_{\epsilon}}$. The outputs of $\mathbf{g_{\epsilon}}$ are then fed into the $\mathsf{Softmax}$ and trained with cross-entropy loss using Adam optimizer \cite{kingma2014adam}.

\subsection{Human-in-the-Loop vs. Conventional Data Splitting Learning}
We investigate I-MPN's abilities to interactively adapt to human feedback provided during the learning model and compare it with a conventional learning paradigm using the fixed train-test splitting rate.

\paragraph{Baseline Setup}
\label{Setup}
In the \textit{conventional machine learning} approach (CML), we employ a fixed partitioning strategy, where the first $70\%$ of video frames, along with their corresponding labels, are utilized for training, while the remaining $30\%$ are reserved for testing purposes. We use I-MPN to learn from these annotations. In the \textit{human-in-the-loop} (HiL) setting, we still utilize I-MPN but with a different approach. Initially, only the first 10 seconds of data are used for training. Subsequently, the model is continuously updated with 10 seconds of human feedback at each iteration. Performance evaluation of both settings is conducted under two scenarios: using the standard testing dataset, with $30\%$ of frames allocated for testing in each video and the whole video. The first one aims to test if the model can generalize to unseen samples, while the latter verifies whether the model suffered from under-fitting.

\paragraph{Result}
Table \ref{tab:cml-vs-iml} showcases our findings, highlighting two key observations. Firstly, I-MPN demonstrates its ability to learn from user feedback, as evidenced by the model's progressively improving performance with each update across various metrics and videos. For example, the mAP$@50_{w}$ score for Video 1 significantly increases from $0.544$ (at $k=0$) to $0.822$ (at $k=2$), reflecting a $51\%$ improvement. Similarly,  Video 2 exhibits a $50\%$ increase in performance, confirming this trend.

Secondly, human-in-the-loop (HiL) learning with I-MPN has demonstrated its ability to match or exceed the performance of conventional learning approaches with just a few updates, even when utilizing a small amount of training samples. For instance, in Videos 1 and 2, after initial training and two to three loops of feedback integration (equating to approximately $18-23\%$ of the total training data), HiL achieves a mAP$@50_{w}$ of $0.835$, while the CML counterpart achieves $0.814$ (trained with $70\%$ of the available data). We argue that such advantages come from user feedback on hard samples, enabling the model to adapt its decision boundaries to areas of ambiguity caused by similar objects or environmental conditions. Conversely, the CML approach treats all training samples equally, potentially resulting in over-fitting to simplistic cases often present in the training data and failing to explicitly learn from challenging samples.

\begin{table}[ht!]
\small
\centering
\setlength{\tabcolsep}{2pt}
\begin{tabular}{c| c c  c  c  c  c  c|  c  c  c c}
\toprule
\textbf{Data} & \textbf{Method} & \textbf{Feedback} & \%\textbf{Data} & \textbf{Time}$_{\text{w}} (s)$ $\downarrow$ & \textbf{mAP}$_{\text{w}}$ $\uparrow$ & \textbf{mAP@50}$_{\text{w}}$ $\uparrow$ & \textbf{mAP@75}$_{\text{w}}$ $\uparrow$ & \textbf{Time}$_{\text{t}}$ (s) $\downarrow$ & \textbf{mAP}$_{\text{t}}$ $\uparrow$ & \textbf{mAP@50}$_{\text{t}}$ $\uparrow$ & \textbf{mAP@75}$_{\text{t}}$ $\uparrow$\\
\midrule
& CML & 0 & 70\% & 401 & 0.66 & 0.814 & 0.771 & 402 & 0.671 & 0.803 & 0.761\\
\cdashline{2-12}
\rule{0pt}{10pt} 
& & 0 & 6\% & 48 & 0.330 & 0.544 & 0.332 & 32 & 0.300 & 0.504 & 0.307\\
Video 1 & HiL & 1 & 6\% & 46 & 0.600 & 0.799 & 0.693 & 29 & 0.541 & 0.732 & 0.656\\
& & 2 & 6\% & 46 & \textbf{0.676} & \textbf{0.822} & \textbf{0.782} & 29 & 0.574 & 0.782 & 0.741\\
& & 3 & 6\% & 46 & \textbf{0.702} & \textbf{0.835} & \textbf{0.793} & 28 & \textbf{0.687} & \textbf{0.809} & \textbf{0.778}\\
\midrule
& CML & 0 & 70\% & 361 & 0.562 & 0.740 & 0.657 & 367 & 0.568 & 0.755 & 0.673\\
\cdashline{2-12}
\rule{0pt}{10pt} 
& & 0 & 5.8\% & 51 & 0.349 & 0.498 & 0.411 & 48 & 0.348 & 0.516 & 0.412\\
Video 2 & HiL & 1 & 5.8\% & 53 & 0.471 & 0.611 & 0.560 & 48 & 0.565 & 0.744 & 0.648\\
& & 2 & 5.8\% & 54 & \underline{0.591} & 0.645 & \underline{0.687} & 48 & \underline{0.581} & \underline{0.762} & 0.662\\
& & 3 & 5.8\% & 54 & \textbf{0.622} & \textbf{0.747} & \textbf{0.683} & 57 & \textbf{0.622} & \textbf{0.800} & \textbf{0.683}\\
\midrule
& CML & 0 & 70\% & 143 & 0.758 & 0.962 & 0.878 & 252 & 0.758 & 0.957 & 0.878\\
\cdashline{2-12}
\rule{0pt}{10pt} 
& & 0 & 8.5\% & 47 & 0.558 & 0.829 & 0.656 & 58 & 0.558 & 0.829 & 0.656\\
Video 3 & HiL & 1 & 8.5\% & 45 & 0.625 & 0.901 & 0.713 & 46 & 0.625 & 0.901 & 0.713\\
& & 2 & 8.5\% & 48 & \textbf{0.764} & \textbf{0.963} & \textbf{0.890} & 57 & \textbf{0.764} & \textbf{0.967} & \textbf{0.880}\\
\bottomrule
\end{tabular}
\caption{Performance comparison between conventional machine learning (\textbf{CML}) and human-in-the-loop (\textbf{HiL}) using I-MPN, evaluated on the \textit{whole video} (\textbf{w}) and evaluated on a \textit{fixed test set} ($30\%$) (\textbf{t}). Feedback $=k$, where $k=0$ indicates the initial training phase, $k > 0$ is the number of times the algorithm is updated. \textbf{Time (s)} is the training time. \textbf{Bold} and \underline{underline} values mark results of \textbf{HiL}, which are higher than \textbf{CML} and represent the best performance overall.}
\label{tab:cml-vs-iml}
\end{table}

\subsection{Comparing with other Interactive Approaches}
In our study, we aim to discriminate the positions of items in the same class, e.g., left and right devices (Figure \ref{fig:environment}). This requires the employed model to be able to explicitly capture spatial relations among object proposals rather than just local region ones. We highlight this characteristic in I-MPN by comparing it with other human-in-the-loop algorithms.

\paragraph{Baselines} (i) The first algorithm we used is the faster-RCNN, which learns from the same human user feedback as I-MPN and generates directly bounding boxes together with corresponding labels for objects in video frames. (ii) The second baseline adapts another deep convolutional neural network (CNN) on top of Faster-RCNN outputs to refine predictions using visual features inside local windows around the area of interest. (iii) Finally, we compare the VoS model used in I-MPN's user annotation collection with the X-mem method~\cite{cheng2022xmem}, but it is now used as an inference tool instead. Specifically, at each update time, X-mem re-initializes segmentation masks and labels, which are given user feedback; then, X-mem propagates these added annotations for subsequent frames. 

\begin{figure}[ht]
    \centering
    \begin{subfigure}[b]{0.5713\linewidth}
        \centering
        \includegraphics[trim={3.5cm .5cm 3cm 2.5cm}, clip, width=\linewidth]{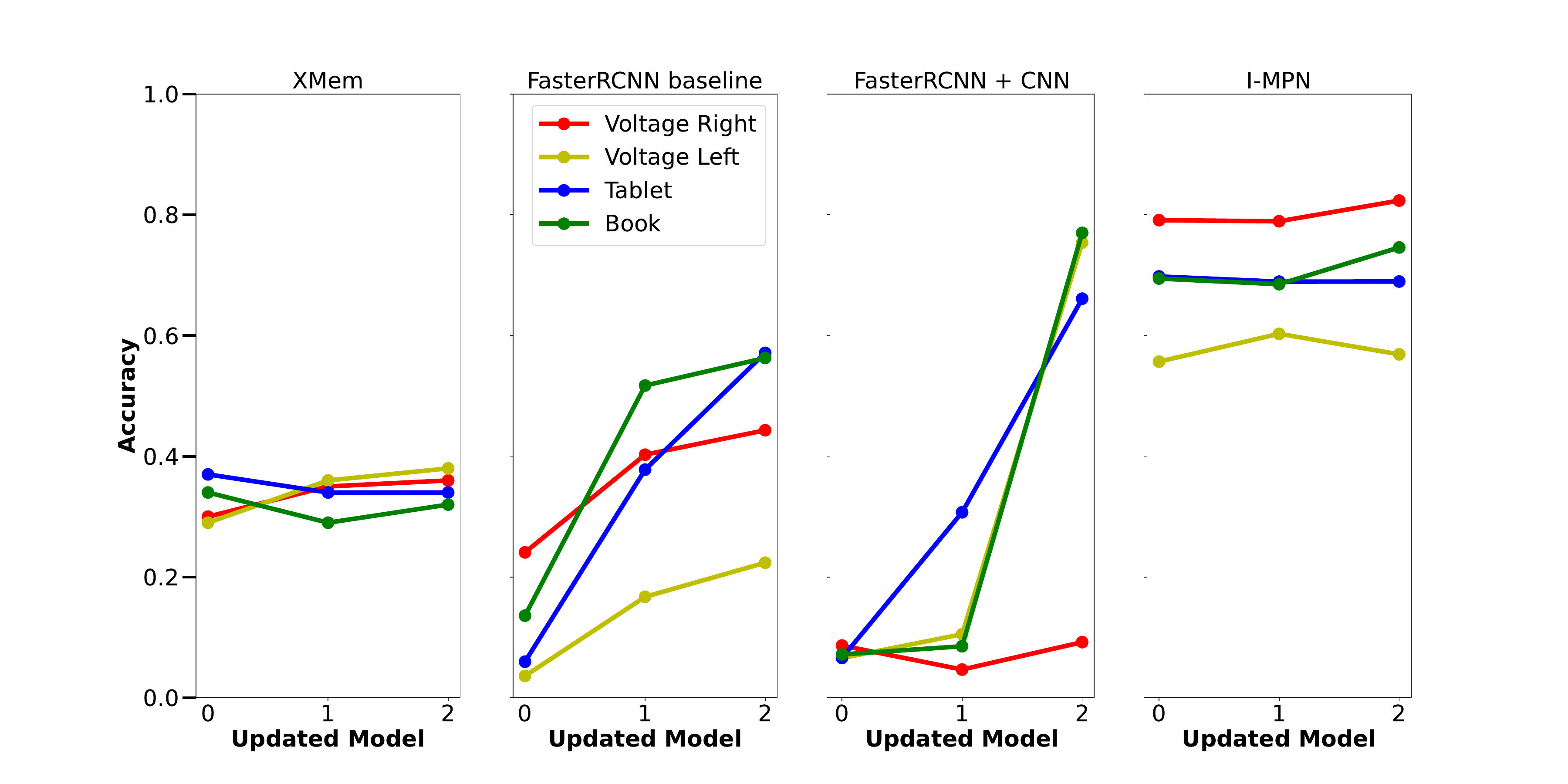}
        \caption{Performance comparison between various human-in-the-loop baselines after each updated time across three video sequences. Results are measured for all objects using the average balanced accuracy metric.}
        \label{fig:iml-compare-all-objects}
    \end{subfigure}
    \hfill
    \begin{subfigure}[b]{0.422\linewidth}
        \centering
        \includegraphics[trim={3.5cm .5cm 3cm 2.5cm},clip, width=\linewidth]{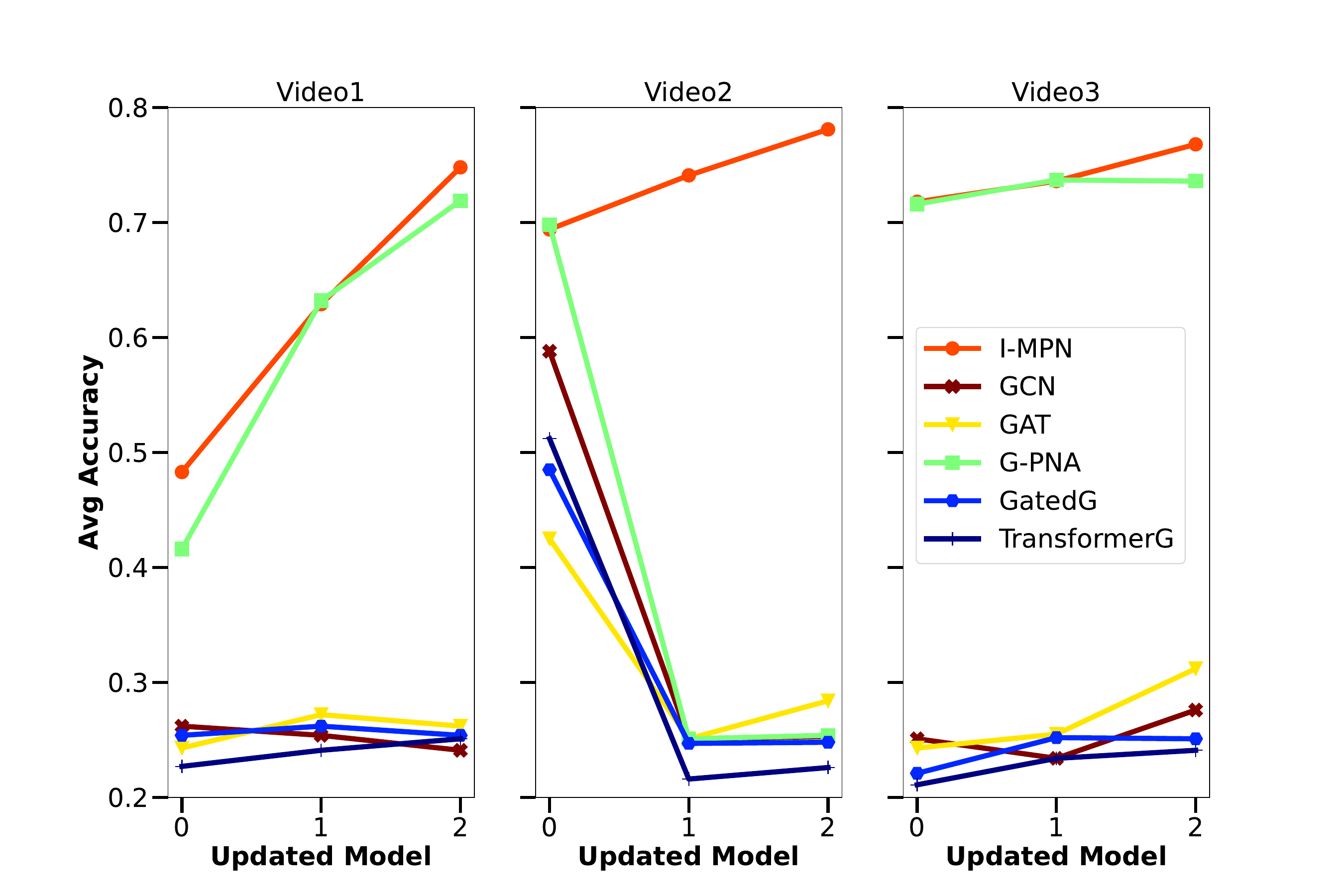}
        \caption{Our \impn\, method uses inductive graph performance compared to other GNNs. Performance is computed for all objects in the $30\%$ test set using average accuracy.}
        \label{fig:gnn_models}
    \end{subfigure}
    \caption{Comparative performance analysis.}
\end{figure}

\paragraph{Results}
We report in Table \ref{tab:iml-baseline-spatial-classes} the performance of all methods in two classes, left and right devices, that require spatial reasoning abilities. A balanced accuracy metric \cite{kelleher2020fundamentals} is used to compute performance at video frames where one of these classes appears and average results across three video sequences. Furthermore, we present in Figure \ref{fig:iml-compare-all-objects} the case where all objects are measured.

It is evident that methods relying on human interaction have consistently improved their performance based on user feedback, except X-Mem, which only re-initializes labels at some time frames and uses them to propagate for the next ones. Among these, I-MPN stably achieved better performance. Furthermore, 
when examining classes such as left and right devices in detail, I-MPN demonstrates markedly superior performance, exhibiting a significant gap compared to alternative approaches. For instance, after two rounds of updates, we achieved an approximate accuracy of $70\%$ with I-MPN, whereas X-mem lagged at only $41.7\%$. This discrepancy highlights the limitations of depending solely on local feature representations, such as those employed in Faster-RCNN or CNN, or on temporal dependencies among objects in sequential frames, like X-mem, for accurate object inference. Objects with similar appearances might have different labels based on their spatial positions. Therefore, utilizing message-passing operations, as done in I-MPN, provides a more effective method for predicting spatial object interactions.


\subsection{Efficient User Annotations}
In this section, we demonstrate the benefits of using video object segmentation to generate video annotations from user feedback introduced in Section \ref{sec:vos}.

\paragraph{Baseline}
(i) We first compare with the \texttt{CVAT} method \cite{sekachev2021cvat}, a tool developed by Intel and an open-source annotation tool for images and videos. CVAT offers diverse annotation options and formats, making it well-suited for many computer vision endeavors, spanning from object detection and instance segmentation to pose estimation tasks. (ii) The second software we evaluate is \texttt{Roboflow}\footnote{\url{https://roboflow.com/}}, another popular platform that includes AI-assisted labeling for bounding boxes, smart polygons, and automatic segmentation. 

\paragraph{Results}
Table \ref{tab:annotation_compare_time} outlines the time demanded by each method to generate ground truth across all frames within three video sequences. Two distinct values are reported: (a) $T_{tot}$, representing the \textit{total} time consumed by each method to produce annotations, encompassing both user-interaction phases and algorithm-supported steps; and (b) $T_{eng}$, indicating the time users \textit{engage} on interactive tasks such as clicking, drawing scribbles or bounding boxes, etc. Notably, actions such as waiting for model inference on subsequent frames are excluded from these calculations.

Observed results show us that using the VoS tool is highly effective in saving annotation time compared to frame-by-frame methods. For instance, in Video 1, CVAT and Roboflow take longer $3$ times than I-MPN on $T_{tot}$. Users also spend less time annotating with I-MPN than other ones, such as $43$ seconds in Video 2 versus $1386$ seconds with Roboflow. We argue that these advantages derive from the algorithm's ability to automatically infer annotations across successive frames using short spatial-temporal correlations and its support for weak annotations like points or scribbles.

\begin{table}[H]
\small
\centering
\setlength{\tabcolsep}{2.5pt}
\begin{tabular}{c |  @{\hspace{0.05in}} c  |@{\hspace{0.05in}}  c | @{\hspace{0.05in}}   c  @{\hspace{0.05in}} c | @{\hspace{0.05in}}  c  @{\hspace{0.05in}} c |  @{\hspace{0.05in}}  c  @{\hspace{0.05in}} c   }
\toprule
 \textbf{Dataset} & \textbf{Time (s)} & \textbf{Frames} &  \multicolumn{2}{c | }{\textbf{Our}}  & \multicolumn{2}{c  | }{\textbf{CVAT}} & \multicolumn{2}{c }{\textbf{Roboflow}} \\
 \midrule
         &           &        & \hspace{0.05in} $T_{tot} \downarrow$      &   $T_{eng} \downarrow$   & \hspace{0.05in} $T_{tot} \downarrow$       & $T_{eng} \downarrow$     &  \hspace{0.05in} $T_{tot} \downarrow$          &   $T_{eng} \downarrow$   \\
          \cmidrule{4-5} 
 \cmidrule{6-7}   \cmidrule{8-9} 
Video 1 & 169 & 3873 & 516  & 74 & 1638  & 1638 & 1722  & 1722 \\ 
Video 2 & 183 & 3422 & 426 & 43 & 1476 & 1476 & 1386 & 1386\\ 
Video 3 & 118 & 2340 & 330 & 36 & 1032 & 1032  & 924 & 924                                        \\
 \bottomrule
\end{tabular}      
\caption{Running time comparison of different methods to generate video annotations. $T_{tot}$ denotes the time taken by each method to infer labels for all frames, while $T_{eng}$ indicates the time users spend actively interacting with the tool through click-and-draw actions, excluding waiting time during mask generation. Smaller is better.}
\label{tab:annotation_compare_time}
\end{table}

\subsection{Further Analysis}
\subsubsection{Inductive Message Passing Network Contribution}
Each frame of the video captures a specific point of view, making the graphs based on these images dynamic. New items may appear, and some may disappear during the process of recognizing and distinguishing objects. This necessitates a spatial reasoning model that quickly adapts to unseen nodes and is robust under missing or occluded scenes. In this section, we demonstrate the advantages of the inductive message-passing network employed in \impn\ and compare it with other approaches.

\paragraph{Baselines} 
We experiment with Graph Convolutional Network (\texttt{GCN})~\cite{kipf2017semi}, Graph Attention Network (\texttt{GAT})~\cite{velikovi2017graph,Brody0Y22}, Principal Neighbourhood Aggregation (\texttt{G-PNA})~\cite{corso2020principal}, Gated Graph Sequence Neural Networks (\texttt{GatedG})~\cite{GatedGCN}, and Graph Transformer (\texttt{TransformerG})~\cite{TransG}. Among these baselines, GCN and GAT employ different mechanisms to aggregate features but still depend on the entire graph structure. G-PNA, GatedG, and Transformer-G can be adapted to unseen nodes, using neighborhood correlation or treating input nodes in the graph as a sequence.

\begin{figure}[ht]

    \centering
    \subfloat[\label{table_fixation_point}]{
        \vspace{.1in}

        \begin{minipage}[b]{0.6\textwidth}
            \centering
            \begin{tabular}{c c c c c}
            \toprule
             \textbf{Video} & \textbf{Object} & \textbf{Initial} & \textbf{Update 1} & \textbf{Update 2} \\ 
             \midrule
              & Avg Acc  & 0.391 & 0.694 & 0.742 \\
              & Voltage & 0.617 & 0.692 & 0.739 \\
             Video 1 & Tablet & 0.274 & 0.912 & 0.966 \\ 
              & Book & 0.189 & 0.350 & 0.489 \\
              & Background & 0.530 & 0.798 & 0.812 \\
             \midrule
              & Avg Acc & 0.501 & 0.755 & 0.839 \\
              & Voltage Left & 0.711 & 0.955 & 0.977 \\
             Video 2 & Tablet & 0.943 & 0.944 & 0.982 \\ 
              & Book & 0.597 & 0.686 & 0.740 \\
              & Background & 0.600 & 0.625 & 0.923 \\
              & Voltage Right & 0.820 & 0.887 & 0.907 \\
             \midrule
              & Avg Acc & 0.250 & 0.726 & 0.748 \\
              & Voltage & 0.182 & 0.222 & 0.667 \\
             Video 3 & Tablet & 0.146 & 0.636 & 0.903 \\ 
              & Book & 0.213 & 0.787 & 0.955 \\
              & Background & 0.766 & 0.851 & 0.971 \\
             \bottomrule
            \end{tabular}
        \end{minipage}
    }
    \hspace{.1in}
    \subfloat[\label{tab:iml-baseline-spatial-classes}]{
       \includegraphics[width=0.24\textwidth]{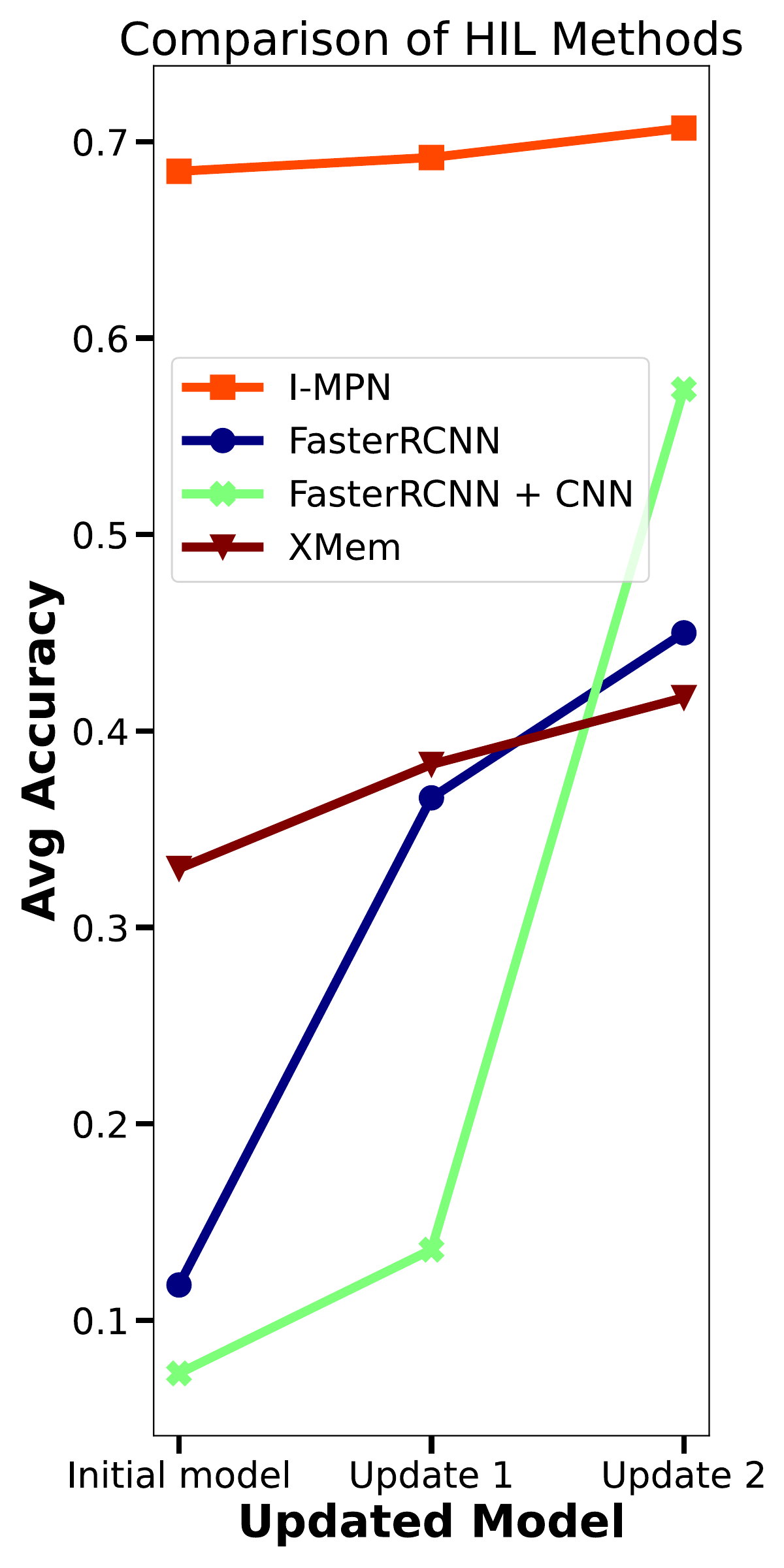}
    }
    \vspace{-.05in}
    \caption{(a) Eye Tracking Point Classification results are improved after upgrading the model with user feedback. Evaluation of different objects given fixation points. (b) Comparison between human-in-the-loop methods on classes requiring spatial object understanding. Results are on balanced accuracy. Higher is better.}
\end{figure}

\paragraph{Results} 
Figure \ref{fig:gnn_models} presents our observations on the averaged accuracy across all objects. We identified two key phenomena. First, methods that utilize the entire graph structure, such as GCN and GAT, struggle to update their model parameters effectively, resulting in minimal improvement or stagnation after the initial training phase. Second, approaches capable of handling arbitrary object sizes, like GatedG and transformers, also exhibit low performance. We attribute this to the necessity of large training datasets to adequately train these models. Additionally, while G-PNA shows promise as an inductive method, its performance is inconsistent across different datasets, likely due to the complex parameter tuning required for its multiple aggregation types. In summary, this ablation study highlights the superiority of our inductive mechanism, which proves to be stable and effective in adapting to new objects or changing environments, particularly in eye-tracking applications.

\subsubsection{Fixation-Point Results}
In eye-tracking experiments, researchers are generally more interested in identifying the specific areas of interest (AOIs) that users focus on at any given moment rather than determining the bounding boxes of all possible AOIs. Therefore, we have further examined the accuracy of our model in the fixation-to-AOI mapping task. Fortunately, this can be solved by leveraging outputs of \impn\, at each frame with bounding boxes and corresponding labels.
In particular, we map the fixation point at each time frame to the bounding box and check if the fixation point intersects with the bounding box to determine if an AOI is fixated (Figure \ref{fig:visualizations}). Similar to our previous experiment, we start with a 10-second annotation phase using the VoS tool after initial training. As soon as there is an incorrect prediction for fixation-to-AOI mapping, we perform an update with a 10-second correction.

\paragraph{Results} Table \ref{table_fixation_point} presents the outcomes of the fixation-point classification accuracy following model updates based on user feedback.
For Video 1, the average accuracy increased from 0.391 at the initial stage to 0.742 after the second update. The classification accuracy for tablets notably increased to 0.966, while books and background objects also exhibited improved accuracies by the second update. For Video 2, an increase in average accuracy from 0.501 to 0.839 was observed. The left voltage object's accuracy reached 0.977, and the right voltage improved to 0.907 by the second update. Tablets maintained high accuracy throughout the updates.
For Video 3, the average accuracy enhanced from 0.250 to 0.748. Tablets and books showed substantial improvements, with final accuracies of 0.903 and 0.955, respectively. The background classification also improved. Overall, the results underscore the effectiveness of user feedback in refining the model's AOI classification, proving the model's adaptability and increased precision in identifying fixated AOIs within eye-tracking experiments.

\subsection{Visualization Results}
\begin{figure}
\centering
    \centering
    \begin{minipage}{\textwidth}
            \centering
            \begin{subfigure}[b]{0.49\linewidth}
                \centering
                \includegraphics[width=\linewidth]{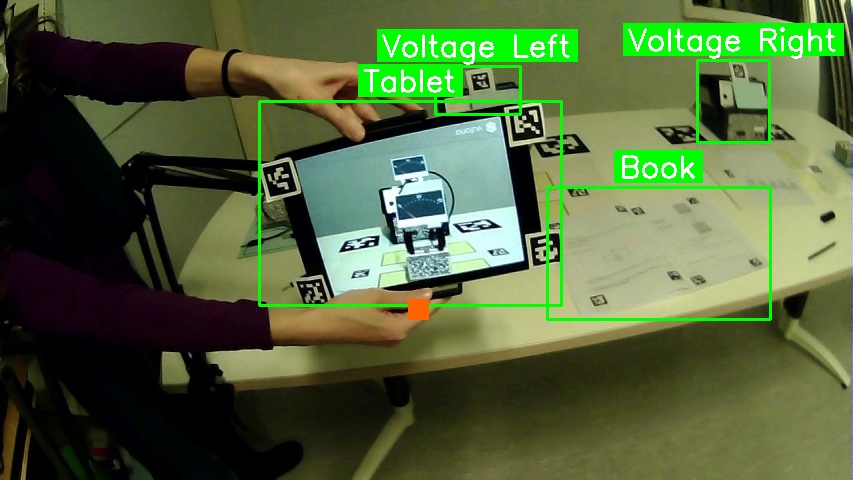}
            \end{subfigure}
            \hfill
            \begin{subfigure}[b]{0.49\linewidth}
                \centering
                \includegraphics[width=\linewidth]{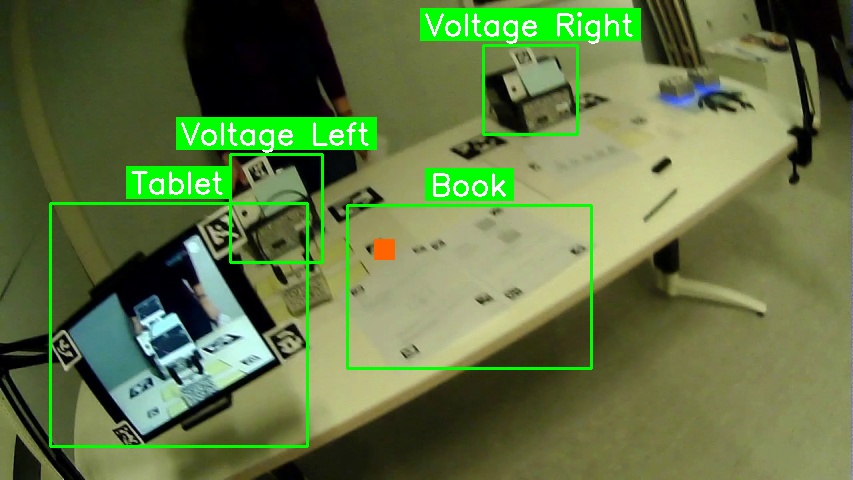}
            \end{subfigure}\\
            \vskip\baselineskip 
            \begin{subfigure}[b]{0.49\linewidth}
                \centering
                \includegraphics[width=\linewidth]{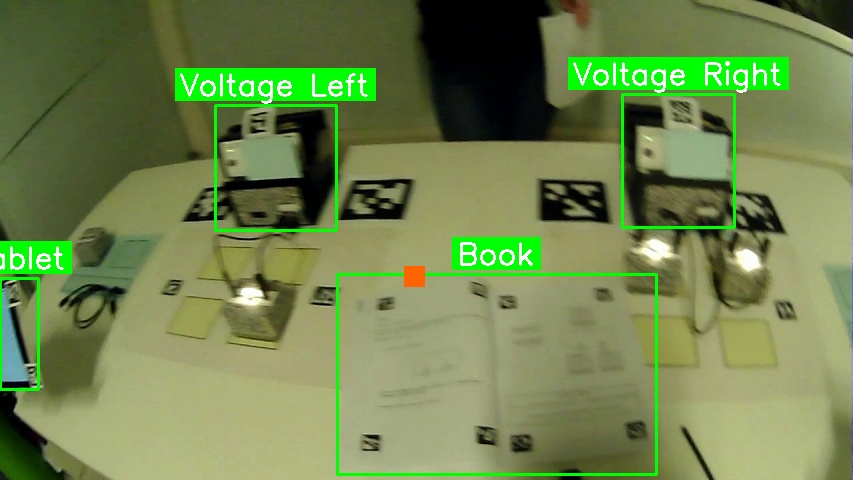}
            \end{subfigure}
            \hfill
            \begin{subfigure}[b]{0.49\linewidth}
                \centering
                \includegraphics[width=\linewidth]{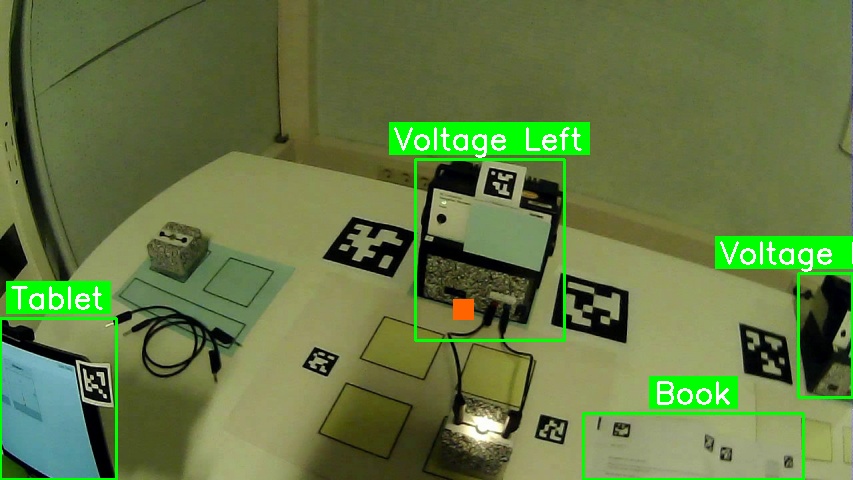}
            \end{subfigure}
        \end{minipage}
    \caption{Visualization results from our interactive-based model, showing fixation points (marked in red) across different video frames.}
    \label{fig:visualizations}
\end{figure}

The visualizations in Figure \ref{fig:visualizations} demonstrate the \impn\, approach's effectiveness in object detection and fixation-to-AOI mapping.
Firstly, even if multiple identical objects are present in a frame, \impn\, is able to recognize and differentiate them and further reason about their spatial location. We see in Figure \ref{fig:visualizations} (bottom left) that both voltage devices are recognized and further differentiated by their spatial location. Additionally, if the objects are only partially in the frame or occluded by another object, \impn\, is still able to recognize the objects reliably. This is especially important in real-world conditions where the scene is very dynamic due to the movements of the person wearing the eye tracker. Lastly, traditional methods that rely only on local information around the fixation point, such as using a crop around the fixation point, can struggle with correctly detecting the fixated object. This is especially true when the fixation point is at the border of the object. This issue is evident in Figure \ref{fig:visualizations}\,(top/bottom left), where traditional methods fail to detect objects accurately. In contrast, our approach uses bounding box information, which allows us to reason more accurately about the fixated AOI. 
In summary, we argue that \impn\, provides a more comprehensive understanding of the scene, particularly in mobile eye-tracking applications where precise AOI identification is essential.

\section{Conclusion and Discussion}
In this paper, we contribute a novel machine-learning framework designed to recognize objects in dynamic human-centered interaction settings. The algorithm is composed of an object detector and another spatial relation-aware reasoning component based on 
the inductive message-passing network mechanism. We show in experiments that our I-MPN framework is proper for learning from user feedback and fast to adapting to unseen objects or moving scenes, which is an obstacle to other approaches. Furthermore, we also employ a video segmentation-based data annotation, allowing users to efficiently provide feedback on video frames, significantly reducing the time compared to traditional semantic segmentation toolboxes.
While \impn\, achieved promising results on our real setups,  we believe the following points are important to investigate:

\begin{itemize}
    \item Firstly, conducting experiments on more complicated human-eye tracking, for example, with advanced driver-assistance systems (ADAS)~\cite{kukkala2018advanced,baldisserotto2023review} to improve safety by understanding the driver’s focus and intentions. Such applications require state-of-the-art models, e.g., foundation models~\cite{zhang2023learning} trained on large-scale data, which can make robust recognition under domain shifts like day and night or different weather conditions. However, fine-tuning such a large model using a few user feedback remains a challenge~\cite{shilong}. 
    \item Secondly, while our simulations using the video object segmentation tool have demonstrated that \impn\ requires minimal user intervention to match or surpass the state-of-the-art performance, future research should prioritize a comprehensive human-centered design experiment. This entails a deeper investigation into how to best utilize the strengths of \impn\ and create an optimal interaction and user interface. The design should be intuitive, minimize errors by clearly highlighting interactive elements, and provide immediate feedback on user actions. These features are important to ensure that eye-tracking data is both accurate and reliable \cite{barz_interactive_2023,jiang2023ueyes}.
    \item Thirdly, extending \impn\, from user to multiple users has several important applications, for e.g., collaborative learning environments to understand how students engage with shared materials, helping educators to optimize group study sessions. Nonetheless, those situations pose challenges related to fairness learning~\cite{yfantidou2023state,shaily2024fairness}, which aims to make the trained algorithm produce equitable decisions without introducing bias toward a group's behavior with several users sharing similar behaviors.
    \item Finally, enabling \impn\, interaction running on edge devices such as smartphones, wearables, and IoT devices is another interesting direction. This 
    ensures that individuals with limited access to high-end technology can still benefit from the convenience and functionality offered by our systems. To tackle this challenge effectively, it is imperative to explore model compression techniques aimed at enhancing efficiency and reducing complexity without sacrificing performance~\cite{marino2023deep,xu2023survey,bolya2022tome,tran2024accelerating}.
\end{itemize}

\bibliography{ref.bbl}

\end{document}